# BERT-Based Combination of Convolutional and Recurrent Neural Network for Indonesian Sentiment Analysis

Hendri Murfi, Syamsyuriani, Theresia Gowandi, Gianinna Ardaneswari, Siti Nurrohmah

Department of Mathematics, Universitas Indonesia, Depok 16424, Indonesia
hendri@ui.ac.id

**Abstract.** Sentiment analysis is the computational study of opinions and emotions expressed in text. Deep learning is a model that is currently producing state-of-the-art in various application domains, including sentiment analysis. Many researchers are using a hybrid approach that combines different deep learning models and has been shown to improve model performance. In sentiment analysis, input in text data is first converted into a numerical representation. The standard method used to obtain a text representation is the fine-tuned embedding method. However, this method does not pay attention to each word's context in the sentence. Therefore, the Bidirectional Encoder Representation from Transformer (BERT) model is used to obtain text representations based on the context and position of words in sentences. This research extends the previous hybrid deep learning using BERT representation for Indonesian sentiment analysis. Our simulation shows that the BERT representation improves the accuracies of all hybrid architectures. The BERT-based LSTM-CNN also reaches slightly better accuracies than other BERT-based hybrid architectures.

## 1 Introduction

The COVID-19 pandemic has forced people to carry out activities from their homes. As a result of this situation, e-commerce platforms have become indispensable for obtaining daily necessities. Bank Indonesia (BI) noted that e-commerce transactions during the pandemic had increased to 547 million transactions with a nominal value of IDR 88 trillion per the first quarter of 2021[1]. Based on a publication from Web Retailer[2], Shopee, Tokopedia, and Lazada are the top three e-commerce platforms with the highest number of visitors per month. The growth of the e-commerce platform is inseparably linked to consumer opinions on its features, both negative and positive. Analyzing the sentiments of reviews on the Google Play Store is one way to monitor the opinions of e-commerce platform customers [1-2].

---

[1] https://www.google.com/amp/s/ekbis.sindonews.com/newsread/472710/39/e-commerce-jadi-andalan-dongkrak-penjualan-di-masa-pandemi-162531223
[2] https://www.webretailer.com/b/online-marketplaces-southeast-asia/

Sentiment analysis, or opinion mining, is a computational study analyzing opinions and emotions expressed in texts [3]. Sentiment analysis can be performed manually, but the more data is used, the more time and effort are required. Therefore, we use machine learning to classify sentiments automatically [4]. Sentiment analysis belongs to supervised learning with target data in sentiment classification, such as positive or negative sentiment. Deep learning is a sub-field of machine learning that has become state-of-the-art in various domains, including sentiment analysis [5]. Convolutional Neural Network (CNN) [6], Long Short-Term Memory (LSTM), and Gated Recurrent Unit (GRU) [7] are three standard deep learning models for sentiment analysis [5].

Previous research has shown that CNN, LSTM, and GRU have excellent performance for sentiment analysis. Next, the researchers tried to improve deep learning performance by combining these standard deep learning architectures. Wang et al. have shown that the hybrid architecture of CNN-LSTM and CNN-GRU perform better than the standard CNN and LSTM alone [8]. They combined the model to take advantage of both the local feature extraction from CNN and long-distance dependencies from RNN. Sosa also proved that the hybrid LSTM-CNN architecture achieved better performance than the standard models and the hybrid CNN-LSTM architecture [9]. Then, Gowandi et al. continue to analyze the performance of the hybrid architectures for Indonesian sentiment analysis in e-commerce reviews [10]. Besides all three hybrid architectures mentioned above, they consider one more hybrid architecture, i.e., GRU-CNN. Their simulations show that almost all hybrid architectures give better accuracy than standard models. Moreover, the hybrid architecture of LSTM-CNN reaches slightly better accuracies than other hybrid architectures.

The model cannot directly process data in text form in sentiment analysis. Hence the data must be transformed into a numeric representation vector. This representation has a significant impact on the model performance. The embedding method is a standard text representation method, a sequence of words. The words are represented by vectors fine-tuned in the learning proses. However, this method ignores the context of words in sentences. Bidirectional Encoder Representation from Transformer (BERT) is the model that has currently become state-of-the-art as a text representation method. This model can construct a representation based on the context of a word in a sentence [5]. Qing, Ziyin, and Kaiwen [11] used BERT to obtain a contextual text representation before conducting sentiment analysis. The results showed that BERT had the best performance compared to various text representation methods.

This paper extends the previous hybrid deep learning by using BERT representation for Indonesian sentiment analysis in e-commerce reviews. The hybrid architectures are CNN-LSTM, LSTM-CNN, CNN-GRU, and GRU-CNN. We compare the performance of the BERT-based hybrid architectures with the previous embedding-based hybrid architectures. Our simulation shows that the BERT representation improves the accuracies of the standard embedding representation for all hybrid architectures. Moreover, the BERT-based LSTM-CNN also reaches slightly better accuracies than other BERT-based hybrid architectures.

The rest of the paper is organized as follows: In Section 2, we present the methods. In section 3, we describe the process of the experiment. In Section 4, we discuss the results of the simulations. Finally, we give a conclusion of this research in Section 5.

## 2   Methods

### 2.1   Convolutional Neural Network (CNN)

A convolutional neural network (CNN) is a deep learning model that has been widely used in text classification problems and has shown to be successful [6]. In-text classification, CNN uses filters to extract essential features from each region. CNN input in word representation will go through two layers, namely the convolution layer and the max-pooling layer [12].

**Convolution Layer.** In the convolution layer, the input will be processed by as many as $l$ filters $W$ to find the essential features of each region with a specific region size. Suppose that the vector representation of the $i$-th word is denoted by $x_i$, and the combination of the vectors of the words $x_i$ to $x_{i+h-1}$ is denoted by $X_{[i:i+h-1]}$. Equation (1) is used to calculate the feature vector $\boldsymbol{c} = [c_1, c_2, \dots, c_{n-h+1}]$ for each filter, where $j$ and $k$ represent the rows and columns of the matrix, and $f$ is a nonlinear activation function. The convolution layer's output is then used as the input for the max-pooling layer.

$$c_i = f\left(\sum_{k=1}^{h} \sum_{j=1}^{d} X_{[i:i+h-1]k,j} \cdot W_{k,j}\right) \tag{1}$$

**Max Pooling Layer.** Max pooling layer processes the output of the convolution layer by taking the most essential features from each feature vector $\boldsymbol{c}$, that is $\hat{c} = \max\{\boldsymbol{c}\}$. This layer aims to reduce the input dimension so that CNN learns to use less information.

### 2.2   Long Short-Term Memory (LSTM)

Long short-term memory (LSTM) is a type of RNN developed by Hochreiter and Schmidhuber [13] in 1997 to remember long-term information. At each training stage (time step), the LSTM has reasonable control over what information should be kept and which should be removed. At the $t$-th time step, LSTM receives two input vectors, that is, the vector representation of the $t$-th word in the sentence ($\boldsymbol{x}_t$) and output vector of the previous hidden state ($\boldsymbol{h}_{t-1}$). The first step in LSTM is to determine what information should be removed from the cell state $\boldsymbol{C}_{t-1}$. This step is determined at the forget gate ($\boldsymbol{f}_t$) shown in equation (2).

$$\boldsymbol{f}_t = \sigma(W_{fx}\boldsymbol{x}_t + W_{fh}\boldsymbol{h}_{t-1} + \boldsymbol{b}_f) \tag{2}$$

The next step is to determine which information will be stored in the cell state $\boldsymbol{C}_t$. This step is divided into two parts: determining the value to be updated through the input gate ($i_t$) as stated in equation (3) and constructing a new vector that is the candidate cell state value ($\widetilde{C}_t$) as shown in equation (4).

$$i_t = \sigma(W_{ix}x_t + W_{ih}h_{t-1} + b_i) \tag{3}$$

$$\widetilde{C}_t = \tanh(W_{cx}x_t + W_{ch}h_{t-1} + b_c) \tag{4}$$

The cell state $C_{t-1}$ is updated to a new cell state ($C_t$) using the outputs of the forget gate, input gate, and candidate vector $\widetilde{C}_t$, as shown in equation (5).

$$C_t = f_t \cdot C_{t-1} + i_t \cdot \widetilde{C}_t \tag{5}$$

The final step is to determine the output using the new cell state. This step is carried out at the output gate ($o_t$) which is shown in equation (6). Then, the vector $o_t$ is used with the cell state $C_t$ to determine the hidden state $h_t$ as in equation (7). The vector $h_t$ will be used in the next time step.

$$o_t = \sigma(W_{ox}x_t + W_{oh}h_{t-1} + b_o) \tag{6}$$

$$h_t = o_t * \tanh(C_t) \tag{7}$$

where $W_{\{fx,fh,ix,ih,cx,ch,ox,oh\}}$ is a weight matrix and $b_{\{f,i,c,o\}}$ is a bias vector [12, 14].

## 2.3 Gated Recurrent Unit (GRU)

Cho et al. proposed a gated recurrent unit (GRU), an RNN variant that can be considered a simplification of the standard LSTM architecture [15]. GRU combines the input gate and forget gate on LSTM into an update gate and adds a gate called the reset gate. Moreover, GRU combines two LSTM states, the cell state, and the hidden state, into a single hidden state [12]. GRU, similar to LSTM, receives input in the form of vector $x_t$ and hidden state $h_{t-1}$ at the $t$-th time step. The first step in GRU is to determine what information needs to be copied from the previous time step. This step is carried out on the update gate ($z_t$) shown in equation (8).

$$z_t = \sigma(W_{zx}x_t + W_{zh}h_{t-1} + b_z) \tag{8}$$

The next step, reset the gate ($r_t$) determines which information from the previous time step should be forgotten. This step is shown in equation (9). The results of the reset gate are then used to determine the hidden state candidate value ($\widetilde{h}_t$), which helps keep crucial old information. The hidden state candidate value is calculated using equation (10).

$$r_t = \sigma(W_{rx}x_t + W_{rh}h_{t-1} + b_r) \tag{9}$$

$$\widetilde{h}_t = \tanh(W_{hx}x_t + r_t \odot W_{hh}h_{t-1} + b_h) \tag{10}$$

The last step is to determine the hidden state $h_t$ which is necessary to store information and transfer it to the next step. Equation (11) is a formula for calculating the hidden state.

$$h_t = z_t \odot h_{t-1} + (1 - z_t) \odot \tilde{h}_t \tag{11}$$

where $W_{\{zx,zh,rx,rh,hx,hh\}}$ is a weight matrix and $b_{\{z,r,h\}}$ is a bias vector [12, 16].

### 2.4 RNN-CNN

The RNN-CNN model is a combined model of the RNN and CNN models. The combined RNN-CNN model architecture refers to both LSTM-CNN and GRU-CNN models. The first step is processing the input with the RNN layer to learn the feature representation and sequence of the data. Then, the output from the RNN layer will be used as input for the CNN layer, which will look for pairs of essential features in the data. Each of the standard models has its advantages. RNN pays attention to word order, while CNN can select important features of ordered word phrases. So, these advantages are expected to maximize the data learning process. Figure 1 shows an illustration of the combined RNN-CNN model's architecture.

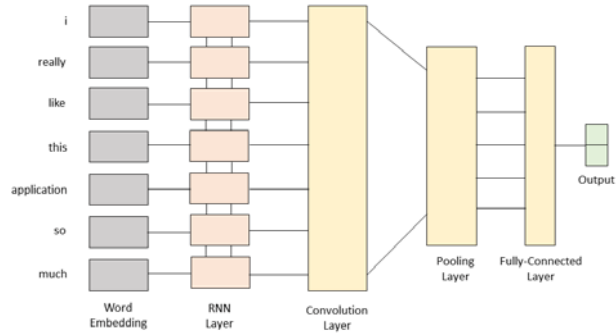

**Fig. 1.** The architecture of the combined RNN-CNN model [9]

### 2.5 CNN-RNN

The hybrid CNN-RNN model has a similar concept to the hybrid RNN-CNN model. The difference lies in the orders of the model. The first step is processing the input with the CNN layer to select word phrases. Then, the output from the CNN layer will be used as input for the RNN layer, which will create a new representation for the data sequence. The combined CNN-RNN model architecture refers to both CNN-LSTM and CNN-GRU models. Figure 2 shows an illustration of the hybrid CNN-RNN model's architecture.

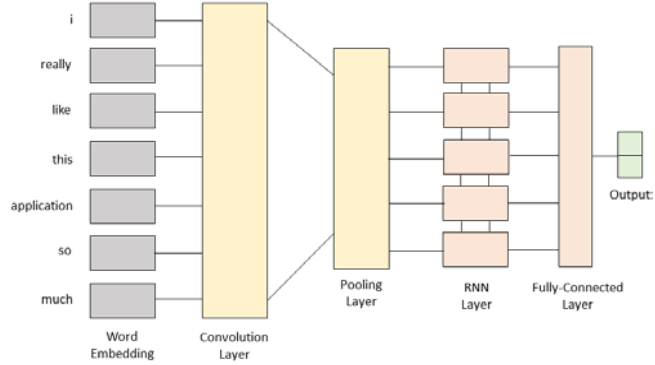

**Fig. 2.** The architecture of the combined CNN-RNN model [9]

### 2.6 Bidirectional Encoder Representation from Transformer (BERT)

BERT is a pre-trained language representation model developed by Devlin et al. [17] in 2018. BERT was developed based on deep learning techniques to improve the capabilities of previous methods like ELMo and OpenAI GPT. BERT is designed to train a bidirectional representation that simultaneously looks at each layer's left and right contexts.

BERT architecture consists of a stack of transformer encoder layers proposed in the research of Vaswani et al. [18]. Each encoder layer consists of two sublayers: multi-head self-attention and position-wise feed-forward network. Multi-head self-attention combines numerous self-attention mechanisms to provide a representation that considers the semantics and connections of each word in the sentence, while a position-wise feed-forward network is used to encode the context for each location separately and identically. The residuals and normalization layers are applied between each of the two sublayers. The BERT model is available in two sizes: $BERT_{BASE}$ and $BERT_{LARGE}$. In this research, we use the $BERT_{BASE}$ model with 12 layers of encoder (transformer blocks), a hidden size of 768, 12 self-attention heads, and 110 million parameters.

To obtain a numerical representation of the text data, we use the feature-based approach of the BERT model. The feature-based method extracts fixed features from a model that has already been trained (pre-trained model). Compared to the fine-tuning approach, the feature-based approach has two advantages. First, different specific model architectures can be added to each task using a feature-based method. This is done because the transformer encoder architecture cannot quickly solve all problems. Second, the feature-based approach improves computational efficiency because the pre-computation representation process is only performed once. Then this representation can be used to perform various experiments [17].

## 2.7 BERT as a text representation method in hybrid deep learning models

This research used hybrid deep learning models, including CNN-LSTM, LSTM-CNN, CNN-GRU, and GRU-CNN. Before performing sentiment analysis with these models, the input in text data is transformed into a numerical representation using BERT.

**BERT-based CNN-(LSTM/GRU).** In CNN-LSTM and CNN-GRU models, the process is passed by each input sentence in Figure 3. The output of the BERT model, a text representation, is first processed by CNN to select the essential features in the data. The CNN output is then sent into the LSTM/GRU model, which creates a new representation by paying attention to the order of the data. Furthermore, a fully connected layer maps the LSTM/GRU output into two classification classes.

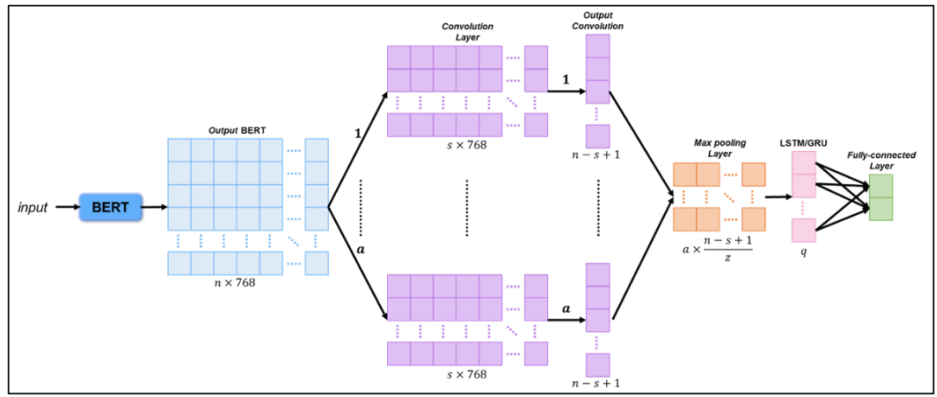

**Fig. 3.** The architecture of BERT-based CNN-LSTM and CNN-GRU

**BERT-based (LSTM/GRU)-CNN.** The architecture of the LSTM-CNN and GRU-CNN models is illustrated in Figure 4. Each input sentence passes through a process similar to the CNN-(LSTM/GRU) model. The difference is that LSTM/GRU receives the BERT output first to learn the feature representation by paying attention to the order of features in the data. The output of LSTM/GRU is then passed into CNN, which looks for essential features in the data.

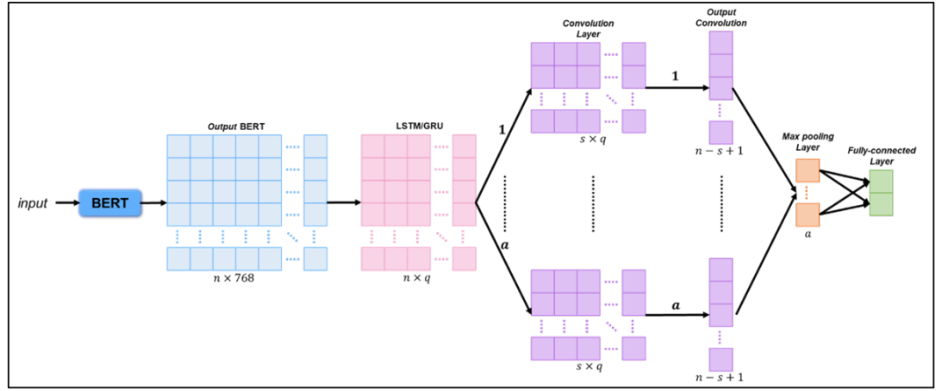

**Fig. 4.** The architecture of BERT-based LSTM-CNN and GRU-CNN

## 3 Experiment

This research begins with data collection, then pre-processing data consisting of data cleaning and one hot encoding. Then, the data is further processed to generate a numerical representation using BERT. The representation results are then sent into the hybrid models CNN-LSTM, LSTM-CNN, CNN-GRU, and GRU-CNN to perform sentiment analysis. Finally, the model performance is evaluated, and the results are analyzed.

### 3.1 Data

The data used is the review of the e-commerce platforms Shopee, Tokopedia, and Lazada from Gowandi, Murfi, and Nurrohmah [10]. Table 1 shows the sentiment description of each data set.

**Table 1.** Description of the data

| Data | Positive Sentiment | Negative Sentiment | Total |
|---|---|---|---|
| Shopee | 2562 | 2435 | 4997 |
| Tokopedia | 1697 | 2663 | 4360 |
| Lazada | 2976 | 3304 | 6280 |

### 3.2 Data pre-processing

Before being used in the model, the input data from each dataset, in-text and sentiment label pairs, are processed. The input data will go through two stages of pre-processing as follows.

1. Data cleaning: there are several data cleaning that is applied to text input, including converting capital letters to lowercase, removing link addresses, removing punctuation and numbers, deleting emoticons, deleting words with only one letter, adjacent letters that are repeated more than twice deleted into only twice and changing the abbreviation using a dictionary from the research of Salsabila, Winatmoko, Septiandri, and Jamal [19].
2. One hot encoding: sentiment labels, which are categorical variables, are converted into numeric variables using one-hot encoding. The data has label 0, which indicates negative sentiment, and label 1, which means positive sentiment. The data label will be mapped into a two-dimensional vector through one-hot encoding. Data with negative sentiment, 0, is mapped to a vector [1,0], while positive sentiment, 1, is mapped to a vector [0,1].

### 3.3 Text representation with BERT

An illustration of the text data representation process with BERT is shown in Figure 3.

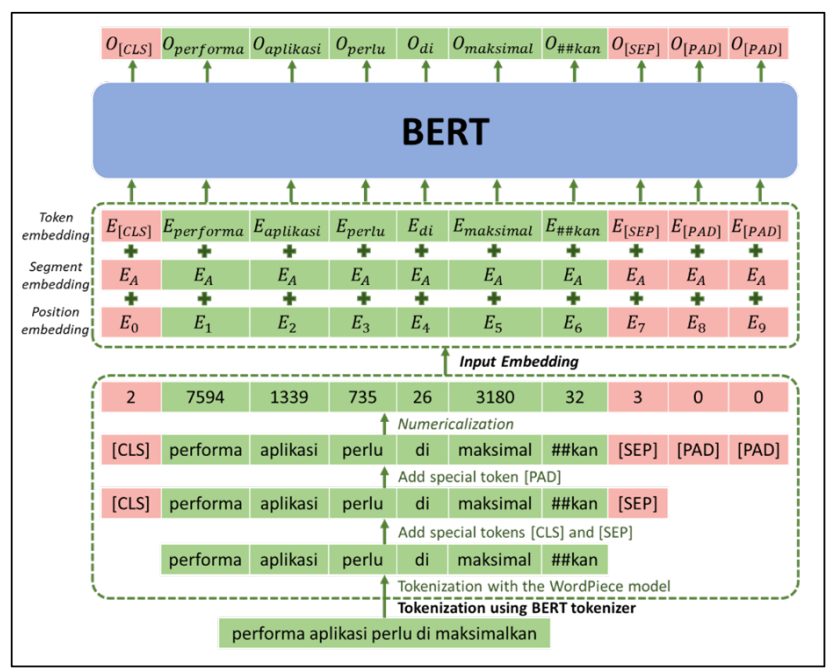

**Fig. 1.** Illustration of text representation with BERT

As illustrated in Figure 3, before BERT processes the text data, it is first adjusted to the input representation that can be accepted by the model. This process consists of several steps, namely tokenization, padding, numericalization, and embedding [20]. BERT tokenizes the input text using the WordPiece model, then adds unique tokens [CLS] at the beginning and [SEP] at the end of the text. After tokenization, padding is required to guarantee that each input sentence has the same length. This study does padding by adding a unique token [PAD] until the sentence length reaches 128 tokens.

Next, each token is transformed into a non-negative integer during the numericalization step. This step is carried out using the WordPiece model vocabulary, consisting

10of 30,522 tokens and unique integers pairs. The numericalization step produces a set of index tokens, then used to map each input token into a numeric vector. This process is carried out at the embedding step, consisting of the sum of the token embeddings, segment embeddings, and position embeddings.

The BERT model used in this study is the IndoBERT$_{BASE}$ proposed by Wilie et al. [21]. This model has 12 encoder layers, 12 attention heads, and a hidden size of 768. Text data representation is obtained by entering the input embedding into the pre-trained BERT model, which is then processed by the encoder layer stack. The first layer encoder will calculate the representation of each token, and the results are used as input to the second layer encoder. This procedure is repeated until the 12th encoder, the last layer encoder. The 12th encoder will produce a contextualized embedding vector representation for each token. The result is a 128×768 matrix, where 128 represents the number of tokens in the text and 768 means the hidden size.

### 3.4 Sentiment analysis simulation

The data in this study is divided into two parts: training data and test data. Each dataset will be split into 80% training data and 20% test data [10]. The training and test data that have passed through the text data representation process using BERT are then used as input to the hybrid models CNN-LSTM, LSTM-CNN, CNN-GRU, and GRU-CNN.

**Table 2.** List of hyperparameter and the candidate values

| Layer | Hyperparameter | Value |
|---|---|---|
| CNN | Number of filters | 200; 250; 300 |
|  | Region size | 5; 4; 5 |
|  | L2 CNN | 0,001; 0,01 |
| LSTM | Unit | 100; 150; 200 |
|  | L2 kernel | 0,001; 0,01 |
|  | L2 recurrent | 0,001; 0,01 |
| GRU | Unit | 100; 150; 200 |
|  | L2 kernel | 0,001; 0,01 |
|  | L2 recurrent | 0,001; 0,01 |
| Fully connected | L2 dense | 0,001; 0,01 |

To build each model, we use candidate hyperparameter values based on research from Gowandi, Murfi, and Nurrohmah [10], shown in Table 2. The model is constructed using 50 epochs, which means that the training data is used 50 times during the learning process. The optimization technique used is Adam with a batch size of 32 and a learning rate of $1 \times 10^{-3}$. Then, the loss function used is categorical cross-entropy. The hyperparameter values in Table 2 will be optimized using Bayesian optimization [22] with a maximum trial of 10 to obtain the optimal combination of hyperparameters. We use early stopping in the model training process with patience five. The value being monitored is validation loss, so if the validation loss increases for five epochs, the learning process will stop early.

Simulations of sentiment analysis for each model were run five times for each dataset. The results of each repetition performance will be averaged and analyzed. The same procedure is also applied to each model with the embedding text representation method and the standard models, but with the addition of an embedding size hyperparameter having candidate values of 64, 100, and 128.

## 4  Results

This section will discuss and analyze the simulation results from single deep learning models (CNN, LSTM, GRU) and hybrid models (CNN-LSTM, LSTM-CNN, CNN-GRU GRU-CNN) with different text representation methods. The sentiment analysis results in each dataset were evaluated using three evaluation metrics: accuracy, precision, and recall. Tables 3, 4, and 5 show the evaluation results of two different text representation methods on other sentiment analysis models using the Shopee, Tokopedia, and Lazada datasets.

Table 3. The evaluation report on the Shopee dataset

| Text Representation Method | Model | Accuracy | Precision | Recall |
|---|---|---|---|---|
| BERT | CNN-LSTM | 0,8572 ± 0,0019 | **0,9058 ± 0,0150** | 0,8058 ± 0,0143 |
|  | LSTM-CNN | 0,8590 ± 0,0023 | 0,8989 ± 0,0201 | 0,8179 ± 0,0201 |
|  | CNN-GRU | **0,8612 ± 0,0054** | 0,9034 ± 0,0188 | 0,8129 ± 0,0190 |
|  | GRU-CNN | 0,8584 ± 0,0023 | 0,8923 ± 0,0127 | **0,8238 ± 0,0177** |
|  | CNN | 0,8492 ± 0,0037 | 0,8920 ± 0,0241 | 0,8047 ± 0,0256 |
|  | LSTM | 0,8526 ± 0,0022 | 0,8975 ± 0,0086 | 0,8047 ± 0,0103 |
|  | GRU | 0,8530 ± 0,0020 | 0,8915 ± 0,0085 | 0,8125 ± 0,0108 |
| Embedding | CNN-LSTM | 0,8344 ± 0,0029 | 0,8882 ± 0,0166 | 0,7754 ± 0,0213 |
|  | LSTM-CNN | 0,8362 ± 0,0022 | 0,8729 ± 0,0088 | 0,7969 ± 0,0076 |
|  | CNN-GRU | 0,8334 ± 0,0030 | 0,8683 ± 0,0152 | 0,7965 ± 0,0137 |
|  | GRU-CNN | 0,8308 ± 0,0033 | 0,8884 ± 0,0231 | 0,7676 ± 0,0239 |
|  | CNN | 0,8195 ± 0,0064 | 0,8671 ± 0,0064 | 0,7957 ± 0,0135 |
|  | LSTM | 0,8396 ± 0,0017 | 0,8869 ± 0,0054 | 0,7879 ± 0,0076 |
|  | GRU | 0,8296 ± 0,0078 | 0,8864 ± 0,0269 | 0,7673 ± 0,0024 |

Table 3 shows the mean and standard deviation of the five trials evaluation of sentiment analysis using different text representations in the Shopee dataset. As shown in Table 3, the model that uses the BERT text representation method produces the best performance, where the model with the best accuracy is CNN-GRU, the model with the highest precision is CNN-LSTM, and the model with the highest recall is GRU-CNN. Furthermore, the text representation method from BERT outperforms the embedding method in this dataset.

**Table 4.** The evaluation report on the Tokopedia dataset

| Text Representation Method | Model | Accuracy | Precision | Recall |
|---|---|---|---|---|
| BERT | CNN-LSTM | 0,8755 ± 0,0021 | 0,8743 ± 0,0128 | **0,7941 ± 0,0131** |
| | LSTM-CNN | **0,8768 ± 0,0033** | **0,8955 ± 0,0235** | 0,7746 ± 0,0249 |
| | CNN-GRU | 0,8764 ± 0,0010 | 0,8944 ± 0,0079 | 0,7735 ± 0,0084 |
| | GRU-CNN | 0,8766 ± 0,0015 | 0,8954 ± 0,0170 | 0,7735 ± 0,0188 |
| | CNN | 0,8681 ± 0,0020 | 0,8764 ± 0,0025 | 0,7693 ± 0,0073 |
| | LSTM | 0,8736 ± 0,0027 | 0,8870 ± 0,0057 | 0,7735 ± 0,0060 |
| | GRU | 0,8677 ± 0,0035 | 0,8735 ± 0,0157 | 0,7711 ± 0,0108 |
| Embedding | CNN-LSTM | 0,8417 ± 0,0011 | 0,8305 ± 0,0065 | 0,7451 ± 0,0088 |
| | LSTM-CNN | 0,8539 ± 0,0021 | 0,8623 ± 0,0138 | 0,7434 ± 0,0201 |
| | CNN-GRU | 0,8525 ± 0,0019 | 0,8718 ± 0,0191 | 0,7286 ± 0,0262 |
| | GRU-CNN | 0,8495 ± 0,0015 | 0,8665 ± 0,0218 | 0,7257 ± 0,0216 |
| | CNN | 0,8537 ± 0,0019 | 0,8637 ± 0,0310 | 0,7386 ± 0,0388 |
| | LSTM | 0,8390 ± 0,0031 | 0,8391 ± 0,0114 | 0,7251 ± 0,0053 |
| | GRU | 0,8294 ± 0,0024 | 0,8197 ± 0,0108 | 0,7198 ± 0,0204 |

Table 4 shows the results of evaluating different text representations on various sentiment analysis models using the Shopee dataset. Based on Table 4, the highest accuracy and precision were achieved by LSTM-CNN, while the highest recall was achieved by CNN-LSTM using text representation from BERT. After that, when compared to the embedding method, the text representation method from BERT produces superior performance.

| Text Representation Method | Model | Accuracy | Precision | Recall |
|---|---|---|---|---|
| BERT | CNN-LSTM | 0,8667 ± 0,0016 | **0,8908 ± 0,0128** | 0,8195 ± 0,0182 |
| | LSTM-CNN | **0,8710 ± 0,0022** | 0,8837 ± 0,0202 | 0,8390 ± 0,0208 |
| | CNN-GRU | 0,8709 ± 0,0017 | 0,8746 ± 0,0103 | **0,8494 ± 0,0125** |
| | GRU-CNN | 0,8669 ± 0,0063 | 0,8867 ± 0,0096 | 0,8300 ± 0,0144 |
| | CNN | 0,8581 ± 0,0021 | 0,8475 ± 0,0090 | 0,8182 ± 0,0144 |
| | LSTM | 0,8683 ± 0,0013 | 0,8744 ± 0,0090 | 0,8434 ± 0,0106 |
| | GRU | 0,8656 ± 0,0010 | 0,8530 ± 0,0050 | 0,8407 ± 0,0074 |
| Embedding | CNN-LSTM | 0,8475 ± 0,0027 | 0,8605 ± 0,0101 | 0,8094 ± 0,0159 |
| | LSTM-CNN | 0,8479 ± 0,0019 | 0,8617 ± 0,0145 | 0,8094 ± 0,0179 |
| | CNN-GRU | 0,8411 ± 0,0032 | 0,8586 ± 0,0099 | 0,7960 ± 0,0195 |
| | GRU-CNN | 0,8455 ± 0,0013 | 0,8597 ± 0,0216 | 0,8067 ± 0,0280 |
| | CNN | 0,8438 ± 0,0025 | 0,8751 ± 0,0147 | 0,8050 ± 0,0212 |
| | LSTM | 0,8468 ± 0,0017 | 0,8603 ± 0,0094 | 0,8081 ± 0,0103 |
| | GRU | 0,8395 ± 0,0014 | 0,8509 ± 0,0054 | 0,8017 ± 0,0078 |

The results of evaluating several sentiment analysis models with different text representation methods in the Lazada dataset are shown in Table 5. From Table 5, it can be seen that the model with the highest accuracy is the LSTM-CNN model, the highest precision is achieved by the CNN-LSTM model, while the model with the highest recall is the GRU-CNN model. These three models use BERT as a text representation method. Similar to the previous two datasets, the BERT text representation method outperformed the embedding method in the Lazada dataset.

Overall, models that used BERT as a text representation method produced the best performance. The BERT-LSTM-CNN model on the Tokopedia dataset had the greatest accuracy of 0,8768. In contrast, the BERT-CNN-LSTM model on the Shopee dataset had the best precision of 0,9058, and the BERT-CNN-GRU model on the Lazada dataset had the best recall 0,8494.

A total of 36 metrics were considered from the three datasets and four hybrid models. The hybrid deep learning models outperformed the CNN model on 21 out of 36 metrics using the embedding method. The hybrid models CNN-LSTM and LSTM-CNN outperformed the LSTM model on 13 out of 18 metrics, while the hybrid models CNN-GRU and GRU-CNN outperformed the GRU model on 16 out of 18 metrics. Meanwhile, the hybrid deep learning models outperformed the CNN model on 35 out of 36 metrics when using BERT as a text representation method. The hybrid models CNN-LSTM and LSTM-CNN beat the LSTM model on 14 out of 18 metrics, and the hybrid models CNN-GRU and GRU-CNN outperformed the GRU model on 17 out of 18 metrics. To show these results, on each dataset, we generated t-SNE visualization of the text data representation using the BERT and embedding method [23].

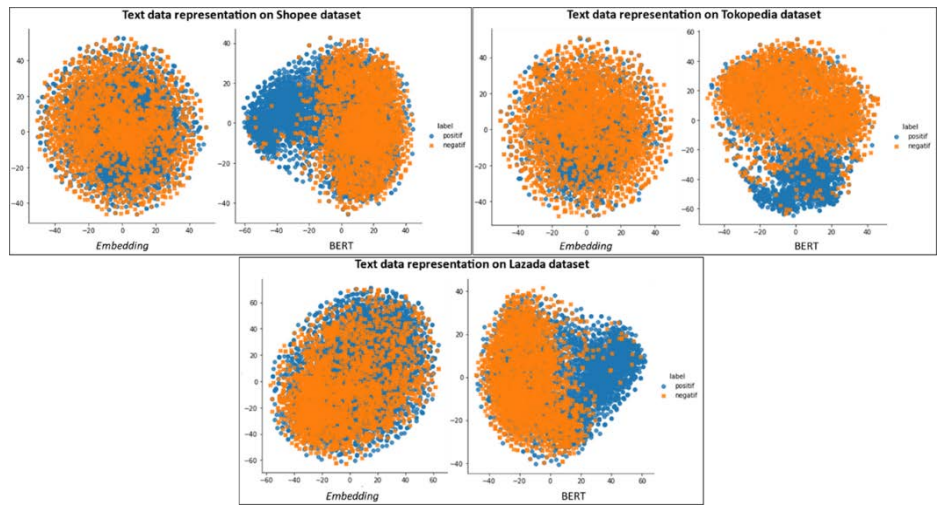

**Fig. 5.** T-SNE visualization of the text representation from BERT and embedding method in each dataset

Figure 5 shows the t-SNE visualization of the text data representation from BERT and the embedding method on each dataset. Based on Figure 4, it can be seen that the sentiment classification from the embedding method for each dataset is not well differentiated. The positive and negative sentiment categories marked in blue and orange are not correctly grouped. In contrast to the embedding method, BERT properly groups positive and negative sentiment classes in each dataset. This is consistent with the model's performance generated using the BERT text representation method, which outperforms the embedding method on each dataset.

## 5  Conclusion

This study used hybrid deep learning models and single deep learning models with two text representation methods, BERT and embedding methods, to analyze sentiment on review data from Indonesian-language e-commerce platforms. The performance of the hybrid models using text representation from BERT was analyzed and compared to the embedding method performance. Furthermore, for each text representation method, the performance of the hybrid models was compared to the performance of the single models. The results showed that by using the BERT text representation method, the best accuracy is achieved by LSTM-CNN on the Tokopedia dataset. The best precision is CNN-LSTM on the Shopee dataset, and CNN-GRU achieves the best recall on the Lazada dataset. Next, the BERT representation outperforms the embedding representation in sentiment analysis using the hybrid architectures CNN-LSTM, LSTM-CNN, CNN-GRU, and GRU-CNN on all datasets. BERT can provide a representation that aligns related texts closer together, which is one of the reasons for its excellent performance in text data representation. Then, on average, hybrid deep learning models deliver superior performance over single deep learning models on all datasets using both text representation methods, BERT, and embedding methods. In the future study, deep learning-based sentiment analysis algorithms can be developed focusing on representation using the BERT model rather than embedding.

## 6  Acknowledgements

Universitas Indonesia supported this paper under a PDUPT 2022 grant. Any opinions, findings, conclusions, and recommendations are the authors' and do not necessarily reflect the sponsors.